\documentclass[letterpaper, 10 pt, conference]{ieeeconf}  

\usepackage[utf8]{inputenc}

\IEEEoverridecommandlockouts
\usepackage{cite}
\usepackage{amsmath,amssymb,amsfonts}
\usepackage{algorithmic}
\usepackage{graphicx}
\usepackage{textcomp}
\usepackage{xcolor}
\usepackage{array}
\usepackage{caption}
\usepackage{graphicx}
\usepackage{float} 
\usepackage{blindtext}
\usepackage{afterpage}
\usepackage{hhline,caption}
\usepackage{hyperref}

\usepackage[utf8]{inputenc}
\usepackage{subcaption}
\usepackage[font=small,labelfont=bf]{caption}
\def\BibTeX{{\rm B\kern-.05em{\sc i\kern-.025em b}\kern-.08em
    T\kern-.1667em\lower.7ex\hbox{E}\kern-.125emX}}

\usepackage{xcolor}
\def\BibTeX{{\rm B\kern-.05em{\sc i\kern-.025em b}\kern-.08em
    T\kern-.1667em\lower.7ex\hbox{E}\kern-.125emX}}

\usepackage[linesnumbered, ruled, longend]{algorithm2e} 

\author{Wen Fan, Max Yang, Yifan Xing, Nathan F. Lepora, Dandan Zhang
}
\begin{document}

\title{\LARGE \bf
Tac-VGNN: A Voronoi Graph Neural Network\\ for Pose-Based Tactile Servoing

\thanks{Wen Fan, Max Yang, Nathan F. Lepora and Dandan Zhang are with the Department of Engineering Mathematics and Bristol Robotics Laboratory, University of Bristol, U.K. Yifan Xing is with the Department of Computer Science, University of Bristol, U.K. Corresponding authors: Dandan Zhang, ye21623@bristol.ac.uk, Nathan Lepora, n.lepora@bristol.ac.uk.}
}

\maketitle
\begin{abstract}
Tactile pose estimation and tactile servoing are fundamental capabilities of robot touch. Reliable and precise pose estimation can be provided by applying deep learning models to high-resolution optical tactile sensors.  Given the recent successes of Graph Neural Network (GNN) and the effectiveness of Voronoi features, we developed a Tactile Voronoi Graph Neural Network (Tac-VGNN)  to achieve reliable pose-based tactile servoing relying on a biomimetic optical tactile sensor (TacTip).  The GNN is well suited to modeling the distribution relationship between shear motions of the tactile markers, while the Voronoi diagram supplements this with area-based tactile features related to contact depth. The experiment results showed that the Tac-VGNN model can help enhance data interpretability during graph generation and model training efficiency significantly than CNN-based methods.
It also improved pose estimation accuracy along vertical depth by 28.57\% over  vanilla GNN without Voronoi features and achieved better performance on the real surface following tasks with smoother robot control trajectories.
 For more project details, please view our website:  \url{https://sites.google.com/view/tac-vgnn/home}
 
\end{abstract}
\setcounter{page}{1}



\section{Introduction}
\label{Introduction}



Tactile perception is needed for robots to  understand the objects they manipulate and the surrounding environment they interact with \cite{luo2017robotic}. Similar to visual servoing where a robot controls the pose of a camera relative to features of the object image, tactile servoing control likewise changes the pose of a tactile sensor in physical contact with an object based on the touch information~\cite{li2013control}.  For example, Fig. 1(a) shows a tactile robotic system  comprising a robot arm (Dobot MG400) and a tactile sensor (TacTip) for surface following, which can be considered a tactile servoing task.  During such a process, the contact between the tactile sensor and the object is changing continuously, through which the surface of the unknown target can be explored.  


Emerging high-resolution optical tactile sensors have been developed and integrated into robotic systems for precise tactile perception, such as GelForce \cite{vlack2005gelforce},  GelSight \cite{yuan2017gelsight}, GelTip \cite{gomes2020geltip}, GelSlim \cite{donlon2018gelslim}, etc. However, their capabilities of sliding over surfaces while maintaining contact have yet to be tested. The most significant difference among different optical tactile sensors is their material properties and construction. GelSight-type sensors have flat sensing surfaces comprised of a molded elastomer. Although highly effective at imaging fine surface details, they are less suited for sliding over surfaces because of the relatively flat and stiff elastomer~\cite{lepora2022digitac}.





\begin{figure}[htbp]
\centerline{\includegraphics[width=3.4in]{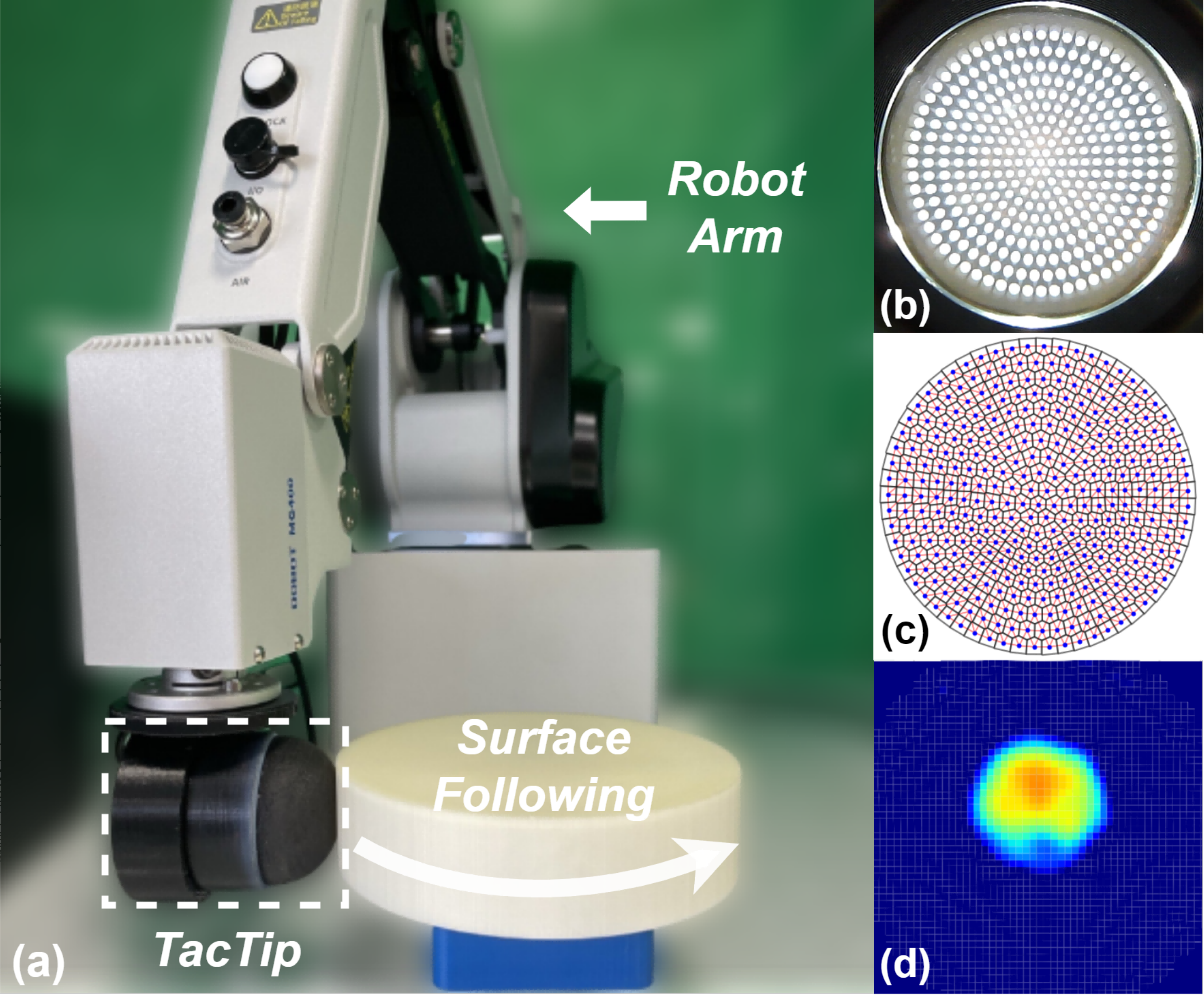}}
\caption{Overview of the experimental setup and the Voronoi Graph for tactile servoing. (a) shows the surface following task conducted by a low-cost  desktop robot arm (Dobot MG400) and a tactile sensor (TacTip). (b) displays the raw image comes from the Tactip sensor. The concept of 'tessellation' can be illustrated in (c). The change of Voronoi enhanced graph data (d) have strong explainability on X, Y, and Z dimension.   }
\vspace{-0.8cm}
\label{intro}
\end{figure}

Among existing low-cost optical tactile sensors, a marker-based optical tactile sensor (BRL TacTip \cite{ward2018tactip}) has shown effectiveness in tactile servoing tasks\cite{lepora2022digitac}. The TacTip has a flexible 3D-printed skin and compliance
from a soft gel, and is therefore 
more practical for the tactile servoing task since it has a greater tolerance for safe contact.  The pins within the TacTip skin are biomimetic in mimicking the dermal papillae in human skin. The key features of contact information can be extracted through computer vision techniques, where deep learning-based methods are playing an increasingly significant role.


 
Convolutional Neural Networks (CNNs) have been widely used to discover latent tactile features for tactile pose estimation \cite{lepora2020optimal}, and have been proven to be effective for pose-based tactile servoing tasks \cite{lepora2021soft, lepora2022digitac}.  However, CNN-based approaches lack interpretability \cite{zhang2021explainable}, since they extract pixel-level features using fixed-size filters and fail to provide explicit information generated from tactile signals. Recently, Graph Neural Networks (GNNs) have gained increasing popularity due to their expressive capabilities and better applicability to non-Euclidean data whose structures are irregular and changeable \cite{zhou2020graph,asif2021graph}. 
The implicit relationship between different markers caused by touch could be well modeled by a GNN through feature-aggregating operations \cite{fan2022graph}. If regarded the pins as vertices, the localization of each pin and the relative distance between any two pins will vary with the skin deformation. This principle is highly similar to the definition of node and edge in a graph, which is a non-linear data structure. Considering that a GNN has a higher efficiency of processing unstructured data, here we explore GNN-based architectures, aiming to construct tactile pose estimation models for pose-based tactile servoing tasks.



The initial step of applying GNNs to the TacTip for pose estimation comes from graph construction.
The K-Nearest Neighbors (kNN) and Minimum Spanning Tree (MST) have been applied to construct graph data \cite{garcia2019tactilegcn,gu2020tactilesgnet}. However, the methods mentioned above are computationally expensive, and thus not applicable to real-time tactile servoing tasks. Meanwhile, Delaunay triangulation has been used to calculate triangulation for a given set of discrete points, which has been demonstrated to have higher robustness compared with kNN~\cite{fortune1995voronoi}. Therefore, Delaunay triangulation will be used to support the graph construction of tactile images in this paper. Moreover, inspired by the study of \cite{cramphorn2018voronoi, gupta2022semi}, the cells generated by the Voronoi tessellation can provide a surrogate of depth information for TacTip sensors. The area change in each cell region should be related to local tactile sensor compression. Therefore, we expect that Voronoi features can be used to enrich the information of graph data converted by tactile images. 
 
 To this end, we define each pin on the TacTip as a node and integrate the pin's position and Voronoi feature as the overall features for the node. The constructed graphs from tactile images are then fed to a GNN model for pose estimation, paving a way for the implementation of tactile servoing. We call this proposed method as Tac-VGNN model, which has advantages over traditional CNN and GNN models in terms of efficiency, interpretability and generalizability.






The \textbf{key contributions} of this paper are listed as follows.
\begin{itemize}
        \item A novel Voronoi Graph representation is designed for processing the tactile information of marker-based optical tactile sensors.
    \item A Tac-VGNN model is developed for tactile pose estimation, while the performances of interpretability and efficiency were evaluated on a surface following task.
\end{itemize}

\section{Methodology}

\subsection{Tactile Graph Generation}

\begin{figure}[htbp]
\centerline{\includegraphics[width=3.4in]{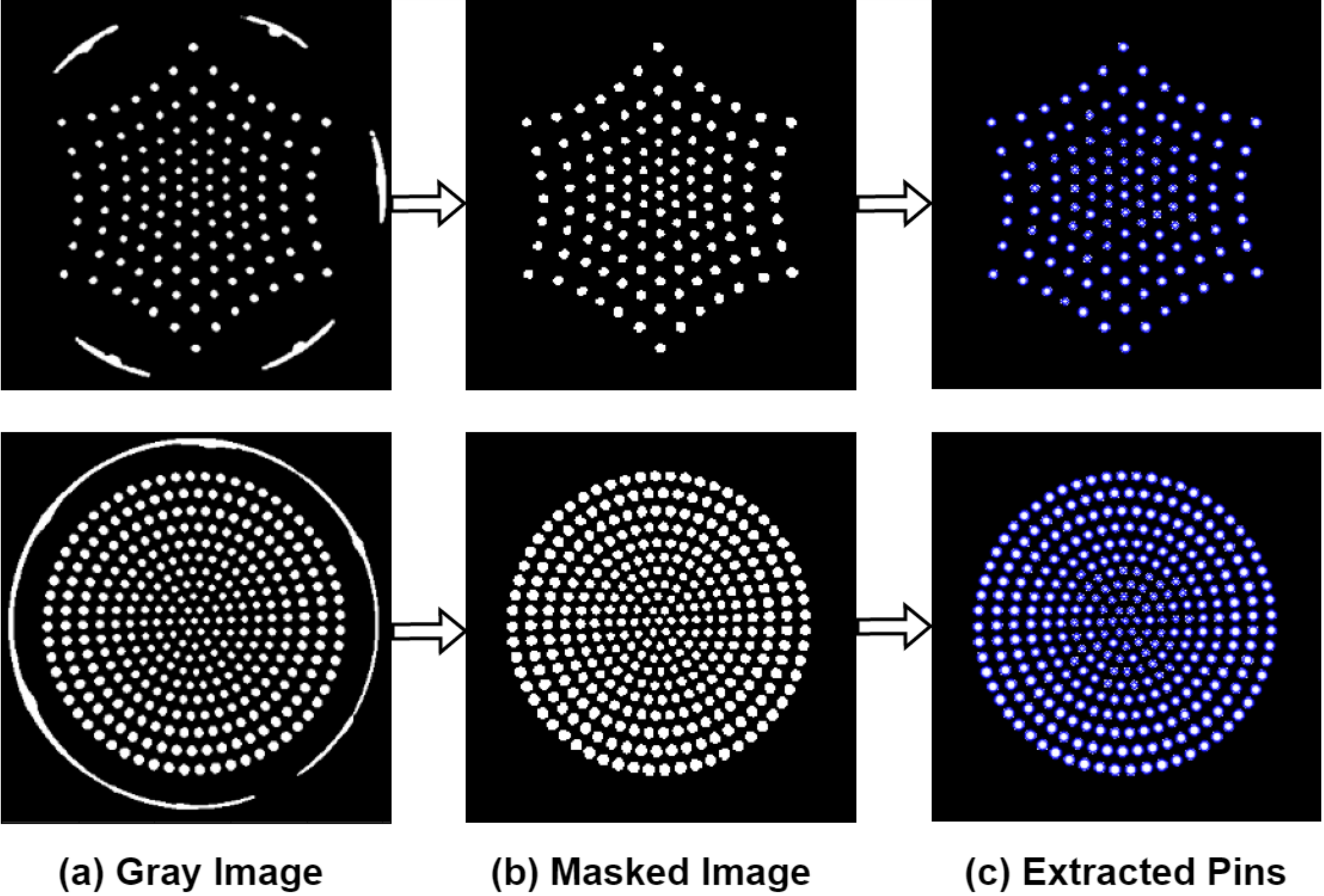}}
\caption{Tactile data preprocessing: (a) the cropped grayscale tactile image; (b) application of circular mask of suitable size to remove artifacts from lighting; (c) pin positions extracted by blob detection.}
\label{preprocess}
\end{figure}

The BRL TacTip can be fabricated at low cost using multi-material 3D printing technology, and customized to different biomimetic morphologies by changing the specifications of sensor diameters, radial depths of skin, and numbers of pins. Two TacTips with different morphologies are selected for experiments in this paper. One TacTip has a hexagonal layout with 127 pins (denoted as `Hexagonal 127'), while the other has a round layout with 331 pins (denoted as `Round 331').
After data collection, the raw tactile images obtained from TacTip are first converted to grayscale images (see Fig.~\ref{preprocess}(a)), then 
erosion and mask filtering are used for noise removal  (see Fig.~\ref{preprocess}(b)).  Subsequently, the positions of pins can be identified via blob recognition (see Fig.~\ref{preprocess}(c)).

\begin{figure}[htbp]
\centerline{\includegraphics[width=3.4in]{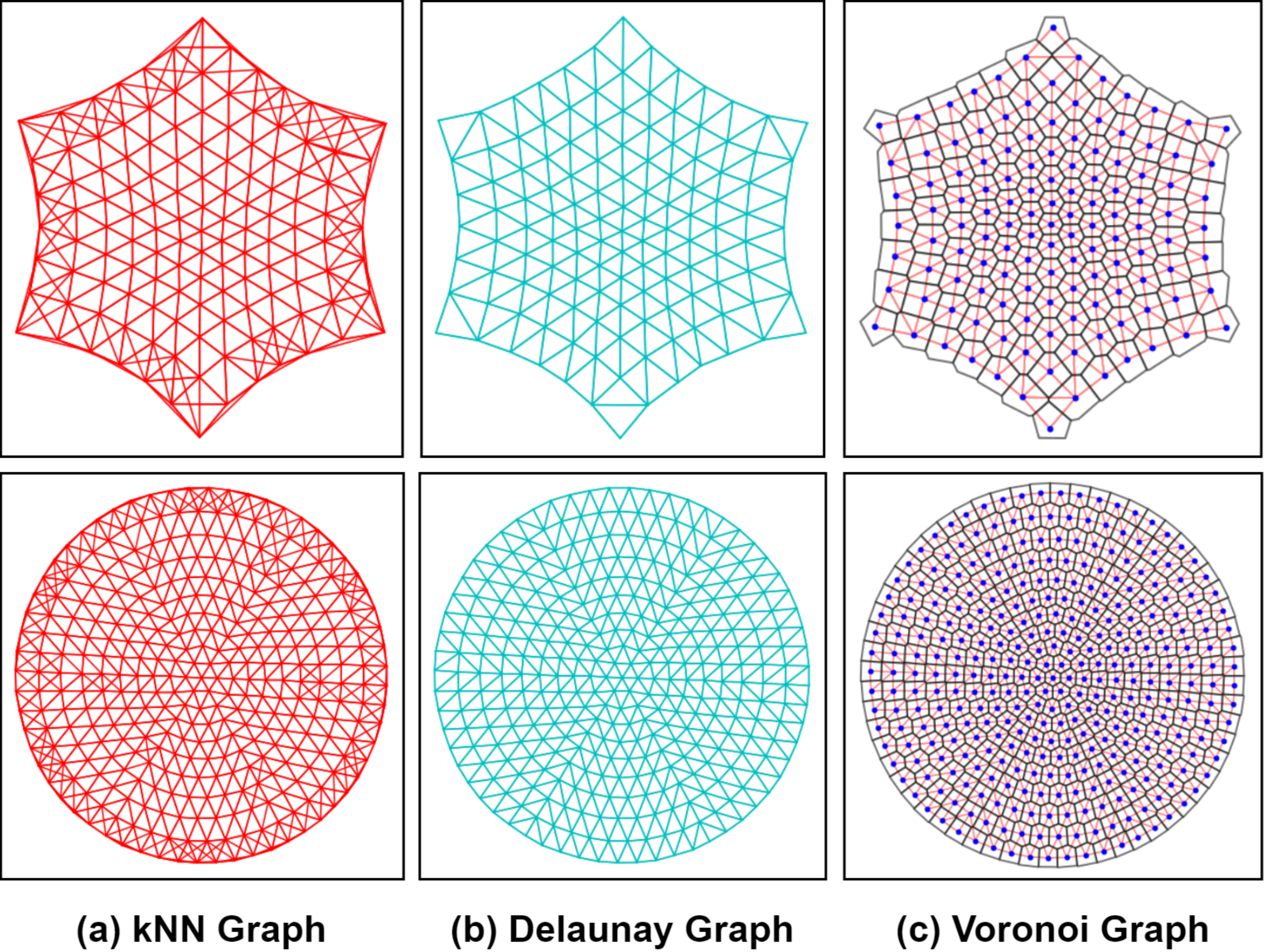}}
\caption{Different graph representations: (a) kNN graph with redundant edges; (b) Delaunay graph; (c) Voronoi graph.}
\label{knn_delaunay}
\vspace{-1em}
\end{figure}


After image preprocessing, two approaches are explored for the tactile graph generation. The kNN approach can be used to build edges between each node and its adjacent nodes; however, redundant edges may be generated for the nodes in the  outermost circle (see Fig.~\ref{knn_delaunay}(a)), which may affect the aggregation performance of the GNN model.
Here, Delaunay triangulation is used to build edges instead of kNN to improve graph building, which generates disjoint triangles for a set of discrete points, as shown  in Fig.~\ref{knn_delaunay}(b). 
Details of the tactile graph generation are given in \textbf{Algorithm~1} from steps 1-5, which build graph nodes $V_i$ and graph edges $E_j$. 
 
 


\subsection{Voronoi Feature Generation}

\begin{figure}[htbp]
\centerline{\includegraphics[width=3.4in]{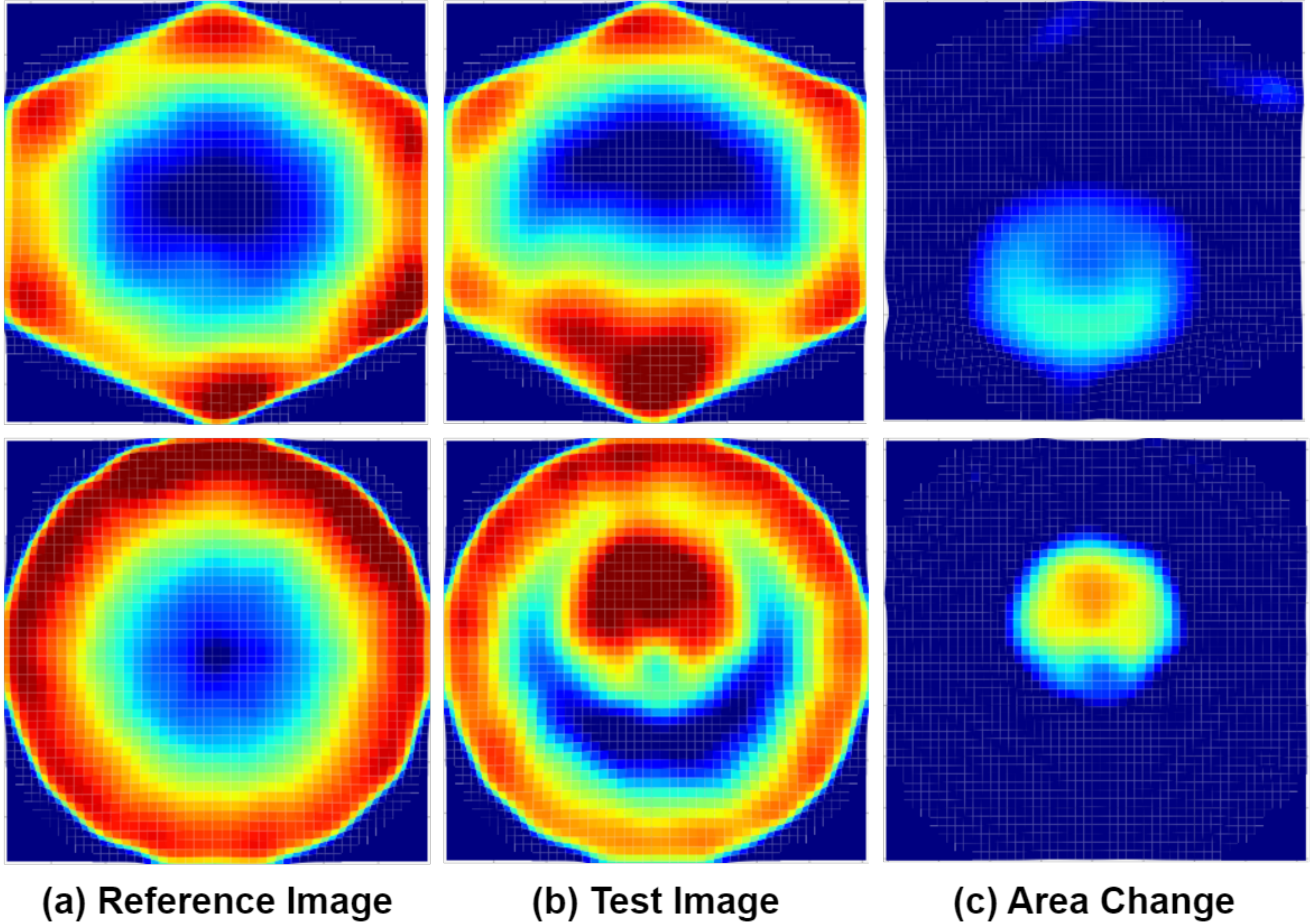}}
\caption{3D plots of Voronoi graph after interpolation operation: (a) shows the reference data which has no contact; (b) represents the Voronoi graph data when deformation occurs; (c) displays the difference between the images in the first and second column.}
\label{voronoi_graph_3d}
\end{figure}

Significant information about the contact distribution and depth of deformation is contained within the shear displacement of pins/markers on the 2D tactile image. The Voronoi tessellation \cite{du1999centroidal} can help extract this valuable information that is highly relevant to the deformation on the sensing surface. \textbf{Algorithm~1}  steps 6-21 describe the details of this procedure. The enlarged boundary $V_{bound}$ is built surrounding $V_i$ to facilitate Voronoi vertice $V_{vertice}$ generation, where non-convex sets, such as Hexagonal 127, need virtual nodes $V_{virtual}$ to assist. The area of each Voronoi tessellation $S_i$ is matched to the location of $V_i$, then combined to form the new node feature $X_i$ = ($V_i$, $S_i$) which relates to the pins' 3D information ($x_i$, $y_i$, $z_i$). Then Voronoi graph can be defined as $G(X_{i}, E_{j})$, as shown in Fig.~\ref{knn_delaunay}(c), providing valuable input for GNN models of sensor pose. 
Fig.~\ref{voronoi_graph_3d} illustrates the 3D plots of Voronoi graph after interpolation. When deformation occurs, the pin density at the contact centre decreases (Fig.~\ref{voronoi_graph_3d}(b)). The difference between two Voronoi graphs can reveal the contact location (Fig.~\ref{voronoi_graph_3d}(c)).

\begin{algorithm}[htbp]
\label{alg3}
 \caption{Process of Voronoi graph generation}
  \KwIn{\emph{$I_{gray}$,} // gray image after cropped and resized}

    $M_{circle} = 255$; // binarized mask\\
    \emph{$I_{processed}$ = AND($I_{gray}$,$M_{circle}$);} // remove noise \\
    \emph{$x_{i}$, $y_{i}$ = Blob($I_{processed}$);} // extracted pins position\\
    \emph{$V_{i}$ = ($x_{i}$, $y_{i}$);} // graph nodes\\
    \emph{$E_{j}$ = Delaunay($V_{i}$);} // graph edges\\
    \emph{$V_{outmost}$ = Convex($V_{i}$);} // outmost nodes\\
    \If{$V_{i}$ is not Convex}
    {
         \emph{$V_{virtual}$ = Rotate($V_{outmost}$);} // virtual nodes\\
         \emph{$V_{convex}$ = $V_{outmost}$ + $V_{virtual}$;} // convex nodes \\
         \emph{$E_{convex}$ = Delaunay($V_{convex}$);} // convex edges\\
         \emph{$E_{j}$ = $E_{convex}$(index($V_{i}$));} // filtered edges\\
         \emph{$V_{outmost}$ = $V_{convex}$;} // new outmost nodes \\
    }

    
    \emph{$V_{bound}$ = $V_{outmost}$ $\times$  $L_{scale}$;} // enlarge boundary \\
    \emph{$V_{new}$ = $V_{bound}$ +  $V_{i}$;} // new nodes set \\
    \emph{$V_{vertice}$, $R_{i}$ = Voronoi($V_{new}$);} // vertices and regions  \\
    \emph{$R_{select}$ = $R_{i}$(index($V_{i}$));} // filtered regions  \\
    \emph{$V_{select}$ = $V_{vertice}$(index($R_{select}$));} // filtered vertices  \\
    \emph{$S_{i}$ = Area($V_{select}$);} // tessellation area size   \\
    \emph{$X_{i}$ = ($V_{i}$, $S_{i}$) ;} // new node feature   \\
    \emph{$G$ = ($X_{i}$, $E_{j}$) ;} // Voronoi graph   \\
      \KwOut{\emph{$G$($X_i$, $E_j$),} // Voronoi-enhanced graph}
\end{algorithm}


To ensure the proposed method can be used for real-time applications, it is important to reduce the computation time for the graph generation. We compare different generation methods using different TacTips, with results summarised in Table~\ref{knn_delaunay_voronoi}. 
One would normally expect the Delaunay graph generation method to have the highest computation speed since it generates fewer edges than the kNN-based method. However, for the hexagonal layout of pins, due to the convex hull computation involved in generating the graph, the computation time is increased; with the round layout of pins, the computation time is lower as is more usual. The time for Voronoi graph generation is the longest but is still within a reasonable range ($<50$\,ms) for real-time operation.


\begin{table}[htbp]
\centering
\caption{Graph Generating Comparison}
\begin{tabular}{ccccc}
\hline
\textbf{TacTip Sensor} & \textbf{Graph Type} & \textbf{Nodes} & \textbf{Edges} & \multicolumn{1}{l}{\textbf{Time Cost}} \\ \hline
Hexagonal 127 pins & kNN      & (127,2) & (762,2)  & 0.004s \\
Hexagonal 127 pins & Delaunay & (127,2) & (744,2)  & 0.007s \\
Hexagonal 127 pins & Voronoi & 
(127,3) & (744,2)  & 0.019s \\
Round 331 pins     & kNN      & (331,2) & (1986,2) & 0.010s  \\
Round 331 pins     & Delaunay & (331,2) & (1860,2) & 0.008s \\
Round 331 pins     & Voronoi & 
(331,3) & (1860,2) & 0.045s \\ \hline
\end{tabular}
\label{knn_delaunay_voronoi}
\end{table}

\subsection{Overview of the Tac-VGNN Model}

\begin{figure}[htbp]
\centerline{\includegraphics[width=3.3in]{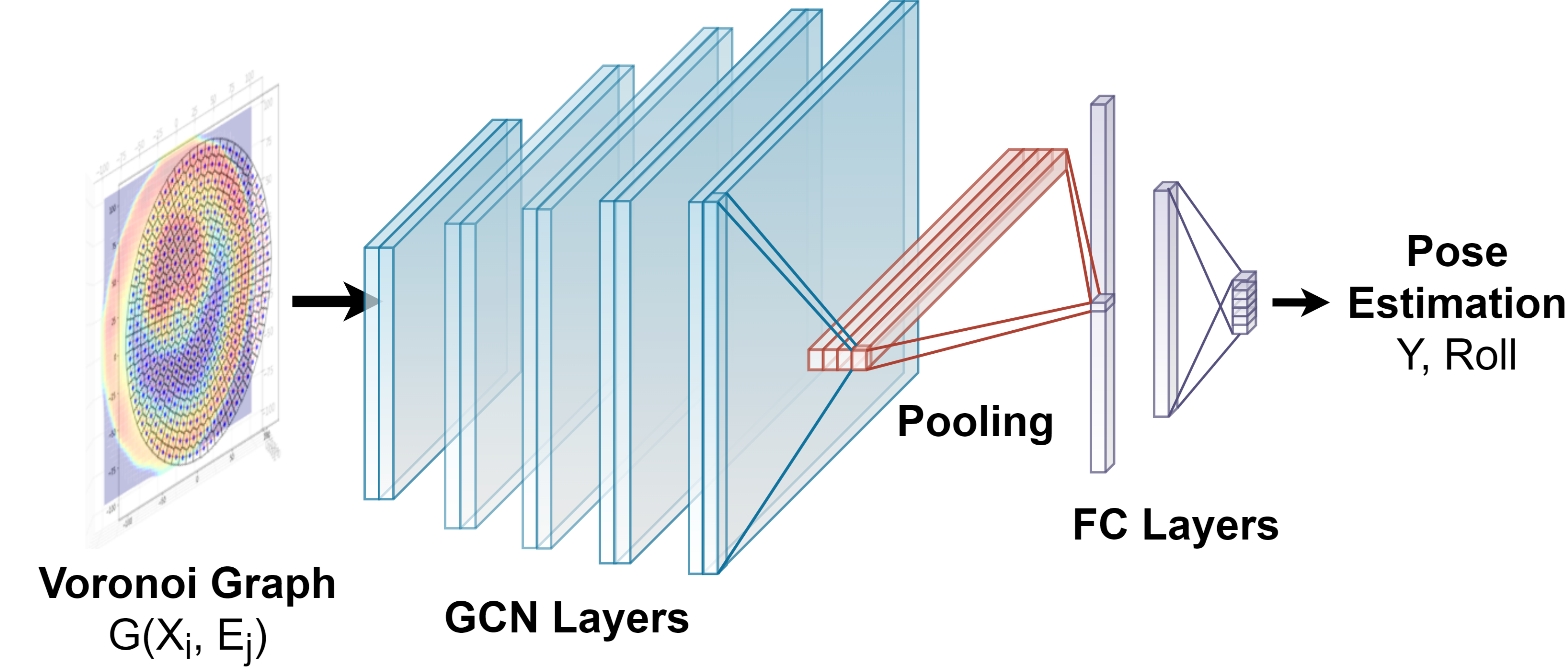}}
\caption{The network architecture of the Tac-VGNN model.}
\label{GNN_framework}
\end{figure}

The Tac-VGNN architecture is shown in Fig.~\ref{GNN_framework}, inspired by Tactile GNN \cite{fan2022graph}. A 5-layer Graph Convolutional Network (GCN) is introduced for feature extraction with filter numbers of (16, 32, 48, 64, 96), whose input is a Voronoi graph $G$($X_i$, $E_j$). A moderate number of GCN layers avoids the risk of over-smoothing while maintaining adequate performance. After the Pooling process, the down-scaled feature vectors are sent to 3 Fully-Connected (FC) layers for pose prediction, whose unit numbers are 96, 64 and 2. The final output consists of the pose estimations for vertical movement $Y$ and rotation angle $\theta_{Roll}$, further explained in Fig.~\ref{2d_pose}(a). 


\section{Experiments and Results Analysis}
\label{Experiments}

\subsection{Tactile Data Collection}



\begin{figure}[htbp]
\centerline{\includegraphics[width=3.4in]{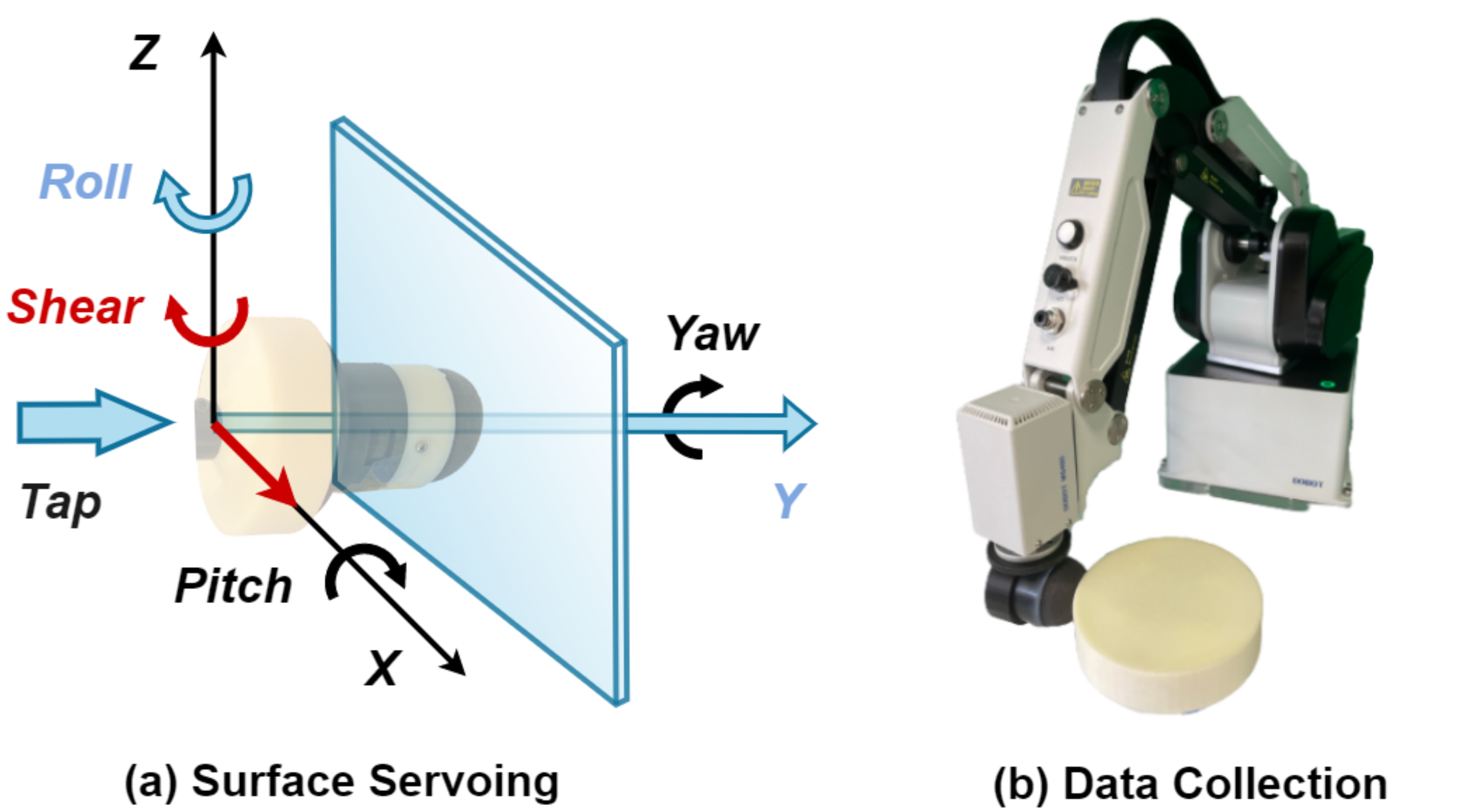}}
\caption{Illustration of pose definition and data collection process. (a) required pose parameter for surface servoing task; (b) an example of data collection using a robot arm. In terms of pose \{$Y$, $\theta_{Roll}$\}, the collection setups have a range of \{[-2, 2]$mm$, [-30, 30]$deg$\}. The tap range is 5$mm$ along $Y$. Shear is added in \{$X$, $\theta_{Roll}$\} with range \{[-5, 5]$mm$, [-5, 5]$deg$\}, facilitating model robustness and generalizability. The number of collected data is 5000.}
\label{2d_pose}
\vspace{-1em}
\end{figure}

During data collection, the TacTip taps repeatedly against a surface with a range of poses augmented with the shear motion to imitate the real contact situations \cite{lepora2019pixels,lepora2020optimal}. The pose of each contact is recorded and saved as labels for each tactile image. The pose parameters for surface tasks include vertical depth $Y$ and rotation angle $\theta_{Roll}$ (see Fig.~\ref{2d_pose}(a)). 
A desktop robot is utilized for the tactile data collection using a TacTip with 331 pins (setup shown in Fig.~\ref{2d_pose}(b)).


\begin{figure}[htbp]
\centerline{\includegraphics[width=3.4in]{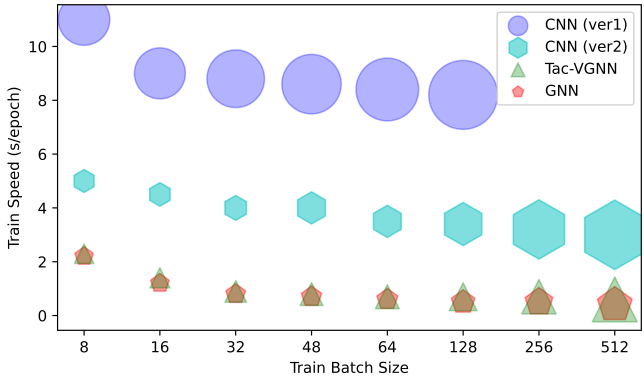}}
\caption{Comparison of model training efficiency. The training time per epoch based on different batch sizes of four models is shown on the diagram. The size of four kinds of markers indicates the GPU memory usage during training for different models. }
\label{efficiency_test}
\vspace{-1em}
\end{figure}


\begin{figure*}[h]
\centerline{\includegraphics[width=6.8in]{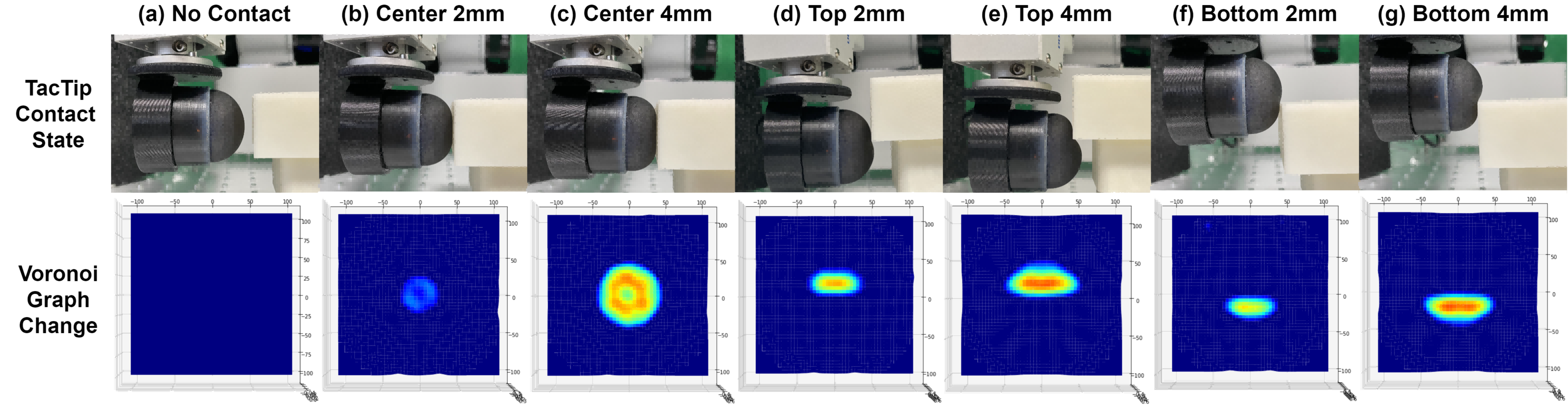}}
\caption{Data interpretability comparison: (a) the reference state without contact, \{$Y$ = 0$mm$, $\theta_{Roll}$ = 0$deg$\}; (b,c) the TacTip central area contacts the surface at different depths, \{$Y$ = [2, 4]$mm$, $\theta_{Roll}$ = 0$deg$\}; (d,e) and (f,g) the TacTip peripheral area touches the object at the top and bottom respectively, \{$Y$ = [2, 4]$mm$, $\theta_{Roll}$ = 0$deg$\}.}
\label{interpretability_experiment}
\end{figure*}

\begin{table}[htbp]\centering
\caption{Comparison of Model Test Performances}
\begin{tabular}{ccccc}
\hline
\textbf{Model}& \textbf{$N_{Conv}$} & \textbf{$N_{FC}$} & \textbf{Pose $Y$ MAE} & \textbf{Pose $\theta_{Roll}$ MAE} \\ \hline
CNN (ver 1)  & 5  & 2  & 0.06 mm   & 1.20 deg  \\
CNN (ver 2)   & 5  & 3 & 0.07 mm   & 1.25 deg  \\
GNN  & 5  & 3  & 0.07 mm   & 1.16 deg  \\
Tac-VGNN  & 5  & 3  & 0.05 mm  & 1.02 deg  \\ \hline
\label{test_table}
\end{tabular}
\vspace{-1em}
\end{table}

\begin{figure}[htbp]
\centerline{\includegraphics[width=3.5in]{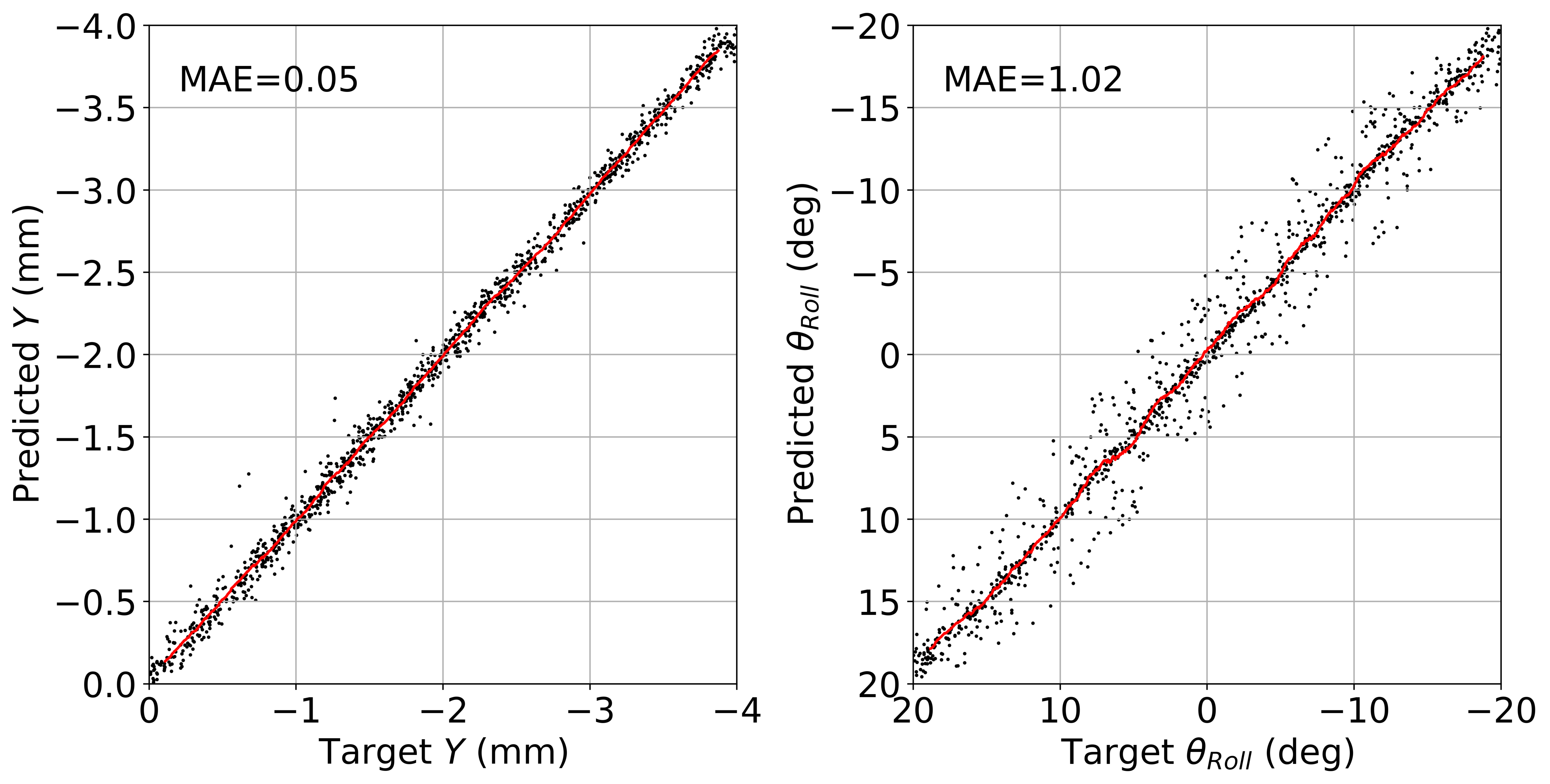}}
\caption{Tac-VGNN model test performance for surface pose estimation, including perpendicular depth $Y$ and rotation angle $\theta_{Roll}$. }
\label{vgnn_test}
\end{figure}

\subsection{Experiment Design}
Four models were used for comparative studies: i) original CNN model (ver 1), ii) customized CNN model (ver 2), iii) GNN model and iv) Tac-VGNN model. Two CNN models work as pose estimation baseline for comparison, whose architecture was designed in \cite{lepora2020optimal, lepora2021pose}, called PoseNet. The CNN (ver 1) has 5 convolutional layers and 2 fully-connected layers, while each convolutional layer has 256 filters. The CNN (ver 2) decreases the filter number for each convolutional layer to 48 and adds an extra fully-connected layer, simulating the regular Tac-VGNN structure. Both GNN and Tac-VGNN have the same architecture but different input dimensions. A kNN-based method is used to generate the graph for GNN model, while a Delaunay-based method is used for Tac-VGNN model. 


To train each model, the tactile dataset of 5000 samples is split randomly into a training/validation dataset (3763 samples; $\sim$75\%) and a test dataset (1237 samples; $\sim$25\%). A Linux PC with Titan XP GPU is used for model training.

\subsection{Offline Analysis}

Here, we conduct comparision studies among 4 models to evaluate their performance in terms of i) Data interpretability; ii) Training efficiency; iii) Prediction accuracy.

\subsubsection{Data Intepretability}

We used different parts of the TacTip to generate deformation at different points of a test object, and let tap depth $Y$ modulate the  deformation level. The images in Fig.~\ref{interpretability_experiment} show the Voronoi graph generated from different levels of deformation. The strong correlation between distribution characteristics and the deformation in reality demonstrates the interpretability of the Voronoi graph. Here, the contact locations are clearly visible and consistent with changes in the density distribution as shown on the heat map. Both the surface and edge contact are clearly visualized. The depth information can also be seen through the color differences between a 2\,mm tap and a 4\,mm tap. In comparison, image data for CNN (Fig.\ref{intro}(b)) and vanilla graphs (Fig.\ref{knn_delaunay}(a)) for GNN have insufficient intuitiveness.

\subsubsection{Training Efficiency}

The training efficiency of each model was evaluated by analyzing its computing time.  As shown in Fig~\ref{efficiency_test}, the CNN (ver 1) was about 10 times slower than the GNNs under all batch sizes. It also has a huge memory footprint, due to more filters being needed within layers, which resulted in GPU memory leaks when the batch size was 256 and 512. The same was also seen with the regular-sized CNN (ver 2) model, which needed more than 2- or 3 times the compute time of GNNs. In contrast, the average training speed of two GNNs were 0.86 and 1.05 sec/epoch respectively, whose 100 epochs of training were completed in less than two minutes. Also, the maximum memory usages were never more than half of the available GPU capacity (5Gb). Evidently, results show both GNNs have significant advantages over CNNs in terms of the training cost. As expected, Tac-VGNN was less computationally efficient than GNN due to extra feature dimensions. We also note that the faster training time of the GNNs will greatly benefit the hyperparameter optimization process, which can be very time-consuming for CNNs.

\subsubsection{Prediction Accuracy}

To compare the prediction capabilities, the test results of each model have been summarised in Table~\ref{test_table}. $N_{Conv}$ and $N_{FC}$ indicate the convolutional and fully-connected layer number. MAE denotes the Mean Absolute Error of pose $Y$ and $\theta_{Roll}$. In all cases, it was found that the rotation angle $\theta_{Roll}$ was harder to predict compared to the vertical pose $Y$, which could be due to the added rotational shear on the Z-axis during the data collection. Overall, the most significant difference was pose $Y$ between  GNN and Tac-VGNN (MAE reduced $28.57\%$), which indicates that the addition of the Voronoi diagram can help learn useful depth-related features. The estimated vertical pose $Y$ determines the contact depth of TacTip, which is closely related to the servoing performance and experiment safety. The test result details of Tac-VGNN are shown in Fig.~\ref{vgnn_test}.





\begin{figure*}[h]
\centerline{\includegraphics[width=6.2in]{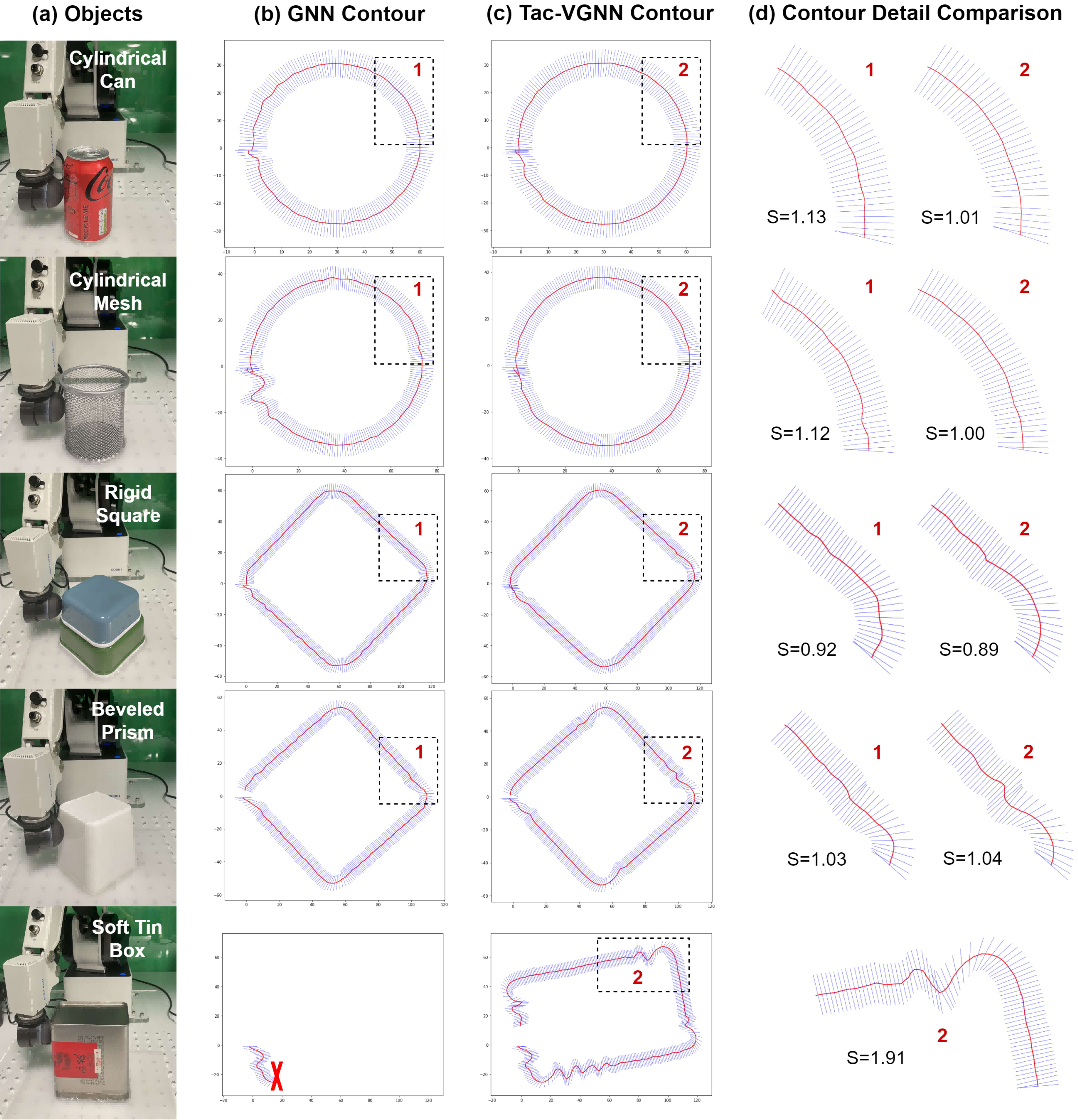}}
\caption{Comparison of surface servoing with different objects: (a) real scene when tactile servoing starts; (b) the trajectories for the GNN model, which fails when servoing around the soft tin box; (c)  the trajectories for the Tac-VGNN model which is the best overall; (d) zoomed-in comparison of trajectories and local smoothness values.}
\label{experiment}
\vspace{-1em}
\end{figure*}


\subsection{Online Tactile Servoing}



To achieve tactile servoing, a Proportional Integral (PI) controller was used to maintain a reference pose designed to keep the TacTip oriented normal to the surface \cite{lepora2019pixels,lepora2021pose}. The input to the controller was received from the model pose predictions. A radial move $\Delta$r and an axial rotation $\Delta\theta$ were generated to move the robot based on the pose error and servo control gain. For the experiment, 5 household items were selected to evaluate the pose models' performance (Fig.~\ref{experiment}(a)). To quantify the tactile servoing performance, we define an evaluation metric called smoothness $S$. By calculating the absolute value of slope between every two points along the trajectory, the average of all these values represents smoothness $S$.
The smaller the value of $S$, the smoother the trajectory. We chose GNN and Tac-VGNN models for comparative studies to focus on the differences between the two GNN-based methods. Five experiments were conducted using five types of objects for surface following, which are detailed as follows.




\subsubsection{Cylindrical Can}

The two pose models gave successful surface following around a cylindrical can, even though the curvature of the surface was significantly different from the flat surface used in training. It appears that the GNN model led to a trajectory with the most fluctuations and the Tac-VGNN model gave a smoother trajectory. This behavior could originate from the differences in the prediction error of $\theta_{Roll}$; it is possible the Tac-VGNN model also generalizes better to curved surface geometries. Smaller cylinders have greater curvature and more pronounced angle changes, which we expect would amplify the differences in model performance during the online surface following task. 

\subsubsection{Cylindrical Mesh}

The addition of a mesh texture onto the cylindrical object did not noticeably degrade the model performance when following the general shape of the surface. Again, the Tac-VGNN gave a smoother trajectory.

\subsubsection{Rigid Square}

The servoing along the straight surfaces of the square objects was the easiest task for all models. Only small deviations were seen at the start of the surface following. The corners were challenging because these were not included in the training data, but both models managed to successfully turn and continue the surface following task.

\subsubsection{Beveled Prism}

The beveled prism was a challenging object because it introduced a tilt relative to the Z-axis that was not present during the data collection and model training (Fig.~\ref{2d_pose}(a)). Both two methods encountered some fluctuations after each corner, which is the most difficult region to predict accurate pose as the controller transitions from circular to linear servoing; however, the robot was able to stabilize and generate smooth trajectories on other parts of the object.

\subsubsection{Tin Box}

The soft tin box was the most challenging as it has flexible surfaces. Any elastic deformation could give a nonlinear change to the tactile sensor that can be misleading for control and bring instability. In Fig.~\ref{experiment}(b), the GNN model without Voronoi enhancement failed when first starting to turn. As the tip moves from the corner to the flatter surface, the elastic deformation can increase drastically, which requires accurate depth predictions to provide responsive control. It further supports the poor estimation of GNN in terms of pose $Y$. In contrast, the Tac-VGNN model performed this task successfully, experiencing transient behavior around corners before reaching steady control.





From our experiments, the Tac-VGNN performed well with fewer fluctuations and more accurate pose estimates compared to the GNN model. Though the Voronoi graph requires a longer generation time, experimental results show that it meets the need for real-time control. The GNN had a worse performance as it lacks reliable depth information and so cannot accurately perceive the deformation, especially during the initial contact, leading to imprecise control of the contact depth as the tactile sensor servos around an object.

\section{Conclusions and Future Work}

In this study, we propose a Tac-VGNN model for tactile pose estimation and evaluated its performance on a surface following task. This method takes full advantage of the additional information provided by Voronoi diagram to characterise the contact depth in each tessellation area. It successfully extends tactile graph features from 2D into 3D contact information, which has advantages in training efficiency, pose prediction accuracy, and increased interpretability. The experimental results indicate that the proposed method has generalizability to complete the surface following tasks of objects with shapes different from those used in training. The interpretability of Tac-VGNN also helps operators build models of objects, such as surface or edge features, facilitating improved dexterous manipulation capabilities. 

Further improvements of Tac-VGNN can be made by leveraging shear or force information to enhance tactile perception.
Additionally, we believe this research could lead to more practical applications, such as medical robots for tactile palpation \cite{zhang2022teleoperation}, service robots for surface cleaning or grasping \cite{zhang2022one},  industrial robots for precise assembly and product quality control \cite{gao2021progress}, and other contact-rich tasks \cite{calandra2018more}.




\bibliographystyle{IEEEtran}

\bibliography{references}

\end{document}